\documentclass[sigconf]{acmart}

\usepackage{xspace}
\usepackage{amsmath}
\usepackage{epsfig}
\usepackage{enumerate}
\usepackage{makecell}
\usepackage{multirow}
\usepackage{epstopdf}
\usepackage{graphicx}
\usepackage{amsfonts}
\usepackage{enumitem}
\usepackage{tikz}
\usepackage{algorithm}
\usepackage{algorithmic}
\usepackage{tabularx}
\usepackage{enumitem}
\usepackage{stmaryrd}
\usepackage{amsthm}
\usepackage{url}
\usepackage{bm}
\usepackage{subfigure}

\theoremstyle{definition}
\newcommand{\mname}{\texttt{COMPOSE}\xspace}

\theoremstyle{definition}
\newtheorem{definition}{Definition}
\newtheorem{problem}{Problem}

\include{def}

\copyrightyear{2020}
\acmYear{2020}
\setcopyright{acmcopyright}\acmConference[KDD '20]{Proceedings of the 26th ACM
SIGKDD Conference on Knowledge Discovery and Data Mining}{August 23--27,
2020}{Virtual Event, CA, USA}
\acmBooktitle{Proceedings of the 26th ACM SIGKDD Conference on Knowledge
Discovery and Data Mining (KDD '20), August 23--27, 2020, Virtual Event, CA, USA}
\acmPrice{15.00}
\acmDOI{10.1145/3394486.3403123}
\acmISBN{978-1-4503-7998-4/20/08}

\begin{document}
\fancyhead{}
\title{\mname: Cross-Modal Pseudo-Siamese Network for \\ Patient Trial Matching}





 \author{Junyi Gao}
\email{junyi.gao@iqvia.com}
\affiliation{%
    \institution{Analytics Center of Excellence}
    \institution{IQVIA}
}

\author{Cao Xiao*}
\email{cao.xiao@iqvia.com}
\affiliation{%
    \institution{Analytics Center of Excellence}
    \institution{IQVIA}
}

\author{Lucas M. Glass}
\email{lucas.glass@iqvia.com}
\affiliation{%
    \institution{Analytics Center of Excellence, IQVIA}
    \institution{Department of Statistics, Temple University}
}

\author{Jimeng Sun}
\email{jimeng@illinois.edu}
\affiliation{%
\institution{Department of Computer Science}
    \institution{University of Illinois Urbana-Champaign}
}

\thanks{*Corresponding author.}
\begin{abstract}

Clinical trials play important roles in drug development but often suffer from expensive, inaccurate and insufficient patient recruitment. The availability of  massive electronic health records (EHR) data and trial eligibility criteria (EC) bring a new opportunity to data driven patient recruitment.  One key task named {\it patient-trial matching} is to find qualified patients for clinical trials given  structured EHR and  unstructured EC  text (both inclusion and exclusion criteria). How to match complex EC text with  longitudinal patient EHRs? How to embed many-to-many relationships between patients and trials? How to explicitly handle the difference between inclusion and exclusion criteria?
In this paper, we proposed  CrOss-Modal PseudO-SiamEse network (\mname) to address these challenges for patient-trial matching. 
One path of the network encodes EC using convolutional highway network. The other path processes EHR  with multi-granularity memory network that encodes structured patient records into multiple levels based on medical ontology. 
Using the EC embedding as query, \mname performs attentional record alignment and thus enables dynamic patient-trial matching. 
\mname also introduces a composite loss term to maximize the similarity between patient records and inclusion criteria while minimize the similarity to the exclusion criteria. Experiment results show  \mname can reach 98.0\% AUC on patient-criteria matching and 83.7\%  accuracy on patient-trial matching, which leads 24.3\% improvement  over the best baseline  on real-world patient-trial matching tasks.

\end{abstract}



\maketitle

\section{Introduction}

Clinical trials with annual market of over 46 billion USD are the only established process for developing new treatments for diseases.  But trials often suffer from expensive, inaccurate and insufficient patient recruitment. 
Many trials struggle to acquire the required number of patients. Moreover 50\% of trials delayed due to patient recruitment issues while some trials are unable to find sufficient patients to begin the trial at all~\cite{risingcosts}. For example, Campbell {\it et al.}~\cite{Campbell07} reported that one-third of publicly funded trials required  time extensions due to insufficient enrollment. It was also reported that $25\%$ of cancer trials failed to enroll  sufficient  patients~\cite{feller15}. Even with sufficient patient, the recruitment cost is high, estimated around 6000 to 7500 USD per patient~\cite{noauthor_undated-ns}.

\begin{figure}[htb!]
\centering
\includegraphics[width=0.8\columnwidth]{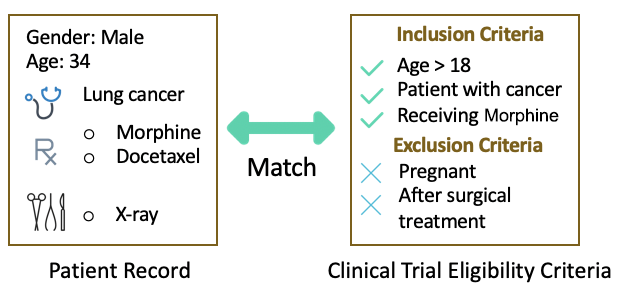}
\vskip -1em
\caption{An illustration of patient-trial matching.}
\label{fig:illustration}
\vskip -1em
\end{figure}

Automated patient-trial matching brings a new opportunity to optimize the trial recruitment process. The key is to find qualified patients for clinical trials given patient's longitudinal electronic health records (EHR) and trial eligibility criteria (EC)  described as both inclusion and exclusion criteria, as shown in Fig.\ref{fig:illustration}. Over the years, rule based retrieval systems and lately a deep embedding  model were developed to match patients for trials. Rule based systems either rely on large amount of human annotations~\cite{weng2011elixr,kang2017eliie} or train supervised learning classifiers to extract rules~\cite{bustos2018learning}, or 
combined machine learning and rule-based methods to generate SQL queries for ECs~\cite{yuan2019criteria2query}. However, they often yield poor recall due to morphological variants and inadequate rule coverage. 
More recently,  DeepEnroll~\cite{zhang2020deepenroll} was proposed as a deep embedding model to predict whether a patient is eligible for a trial. It jointly embeds and aligns medical concepts derived from patient data and EC onto the same embedding space, and then perform simple similarity match of patients to trials based on the aligned embeddings. However, DeepEnroll ignores the difference between inclusion and exclusion criteria, which can lead to criteria mismatch. 
In general, here are the key open challenges for patient trial matching:

\begin{enumerate}[leftmargin=*]

    \item \textbf{Multi-granularity medical concept}.  Unstructured ECs often encode more general disease concepts due to heterogeneity of disease manifestation~\cite{hill08}, while structured patient EHR often represent patient conditions using more specific medical codes. For example, a patient with \textit{pleuropericardial adhesion} on EHR data can be recruited to a trial that require more general \textit{cardiovascular diseases}. It is nontrivial to match medical concepts with heterogeneous granularity in different data modalities.
    \item \textbf{Many-to-many relationship between patients and trials}. In practice, each patient may enroll more than one trial and vice versa. Since each trial generally focus on certain diseases, the semantic distance between trials even ECs may be large. However, all existing works align the patient embedding to different trial embeddings, which may confuse the embed function and lead to inferior matching results.
   
    \item \textbf{Explicit Inclusion/Exclusion Criteria Handling}. Trial ECs comprise  inclusion and exclusion criteria. They describe what are desired and unwanted  from the targeted patients. Existing approaches did not explicitly differentiate these two criteria, which may significantly affect the  matching accuracy.
\end{enumerate}

\noindent
To address these challenges, we propose CrOss-Modal Pseudo-Siamese (\mname) for patient trial matching, which has the following contributions:

\begin{enumerate}[leftmargin=*]
    \item \textbf{Taxonomies guided multi-granularity medical concept embedding}. To match  medical concepts of heterogeneous granularity, we augment the medical codes in patient's records with their 
    textual descriptions and hierarchical taxonomies, such that concepts can be embedded in both finer and more coarse levels to better align detailed medical codes in EHR and medical concepts with various granularity.
    \item \textbf{Attentional record align and dynamic patient-trial match}.
    Instead of aligning a patient's entire record to each trial, we developed an attentive READ mechanism inside of a dynamic memory network to extract the best matching part of patient EHR to match with ECs at criteria level.
    \item \textbf{Differentiating inclusion/exclusion criteria}.  \mname also 
    has a composite similarity loss term to explicitly handle the inclusion and exclusion criteria separately. It improved the patient-trial matching based on maximizing the similarity between patient and inclusion criteria while minimizing the patient and exclusion criteria in latent space.
\end{enumerate}

We evaluated \mname on real world clinical trial dataset. \mname significantly outperformed the best state-of-the-art baselines. It achieved 24.3\% relatively higher accuracy over the best baselines on patient-trial matching tasks.

\section{Related Works}
\textbf{Patient trial matching} can be categorized as
rule based  systems and deep embedding based models. Rule based systems try to extract named entities and relations for trial eligibility criteria (ECs) and construct rules for identifying patients. They either rely on large amount of human annotations~\cite{weng2011elixr,kang2017eliie} or train supervised learning classifiers to extract rules~\cite{bustos2018learning}, or 
combined machine learning and rule-based methods ~\cite{yuan2019criteria2query} to rules for ECs. For example, EliXR~\cite{weng2011elixr} matches Unified Medical Language System (UMLS) concepts and relations via pre-defined dictionary and regular expressions.
Alicante \textit{et al.}~\cite{alicante2016unsupervised} utilized unsupervised clustering methods for eligible rule extraction. Bustos \textit{et al.} ~\cite{bustos2018learning} used naive machine learning models such as CNN, SVM, and kNN as classifiers for specific disease EC classification. Yuan \textit{et al.} proposed a complete pipeline Criteria2Query~\cite{yuan2019criteria2query}, which combines machine learning and rule-based methods to form patient criteria. 
These methods often yield poor recall due to  morphological variants and inadequate rule coverage. 
Recent years, deep embedding based models such as DeepEnroll~\cite{zhang2020deepenroll} jointly embeds patient records and trial ECs in the same latent space, and then aligns them using attentive inference. However, DeepEnroll did not consider the match of different concept granularity nor differentiate inclusion and exclusion criteria.  In the experiment, we compare with Criteria2Query and DeepEnroll as they are the state of the art methods.  \\

\noindent\textbf{Cross-modal retrieval} enables flexible retrieval across different modalities, such as semantic image-text retrieval \cite{you2018end, ren2015multi}. The core of cross-modal retrieval is how to measure the content similarity. Among others, Siamese network is a typical structure for uni-modal retrieval tasks~\cite{guo2017learning,qi2016sketch}. It consists of two branches of the same structure and utilizes similar pairs and dissimilar pairs for similarity learning. Pseudo-Siamese network~\cite{hughes2018identifying,treible2019wildcat} is more flexible than Siamese network in the sense that it allows different structures to receive inputs from different modalities. In this work, we developed a pseudo-siamese network for patient trial matching task. 
\begin{figure*}[htb!]
\centering
\includegraphics[width=1.6\columnwidth]{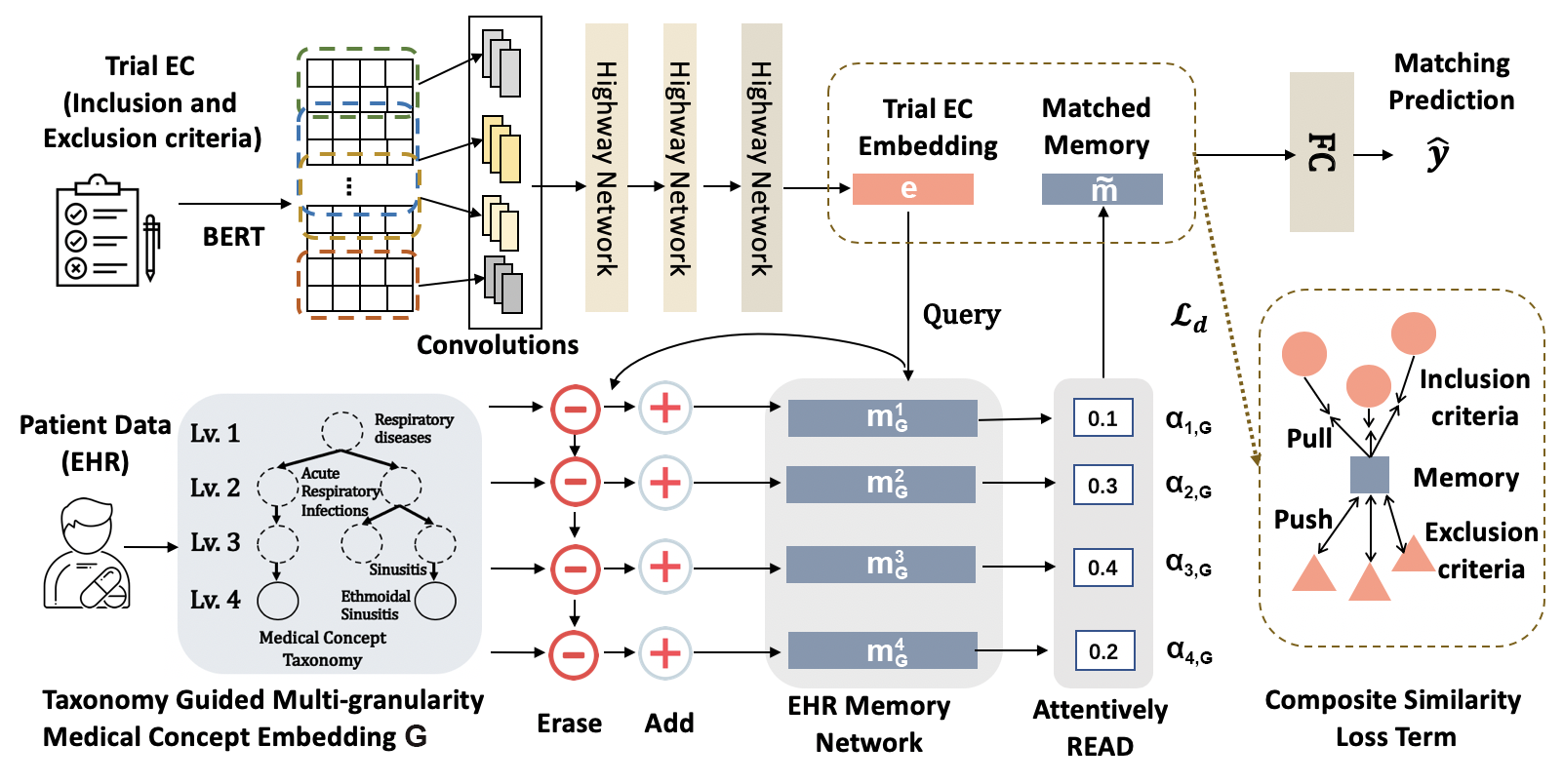}
\vskip -1em
\caption{\mname model. (1) Clinical trial EC sentence embedding: First, we use pretrained BERT to generate contextualized word embeddings for each word in EC sentences. Then we use CNNs to capture semantic features and generate embeddings for EC sentence. (2) Taxonomy guided multi-granularity medical concept embedding: We use hierarchical memory networks to store medical concept taxonomies. Different granularity information of concept is stored in different levels of the memory network. (3) Attentional record align and dynamic patient-trial match: We use learned EC embeddings to attentively visit the memory network and retrieve the best matching memories $\widetilde{\bm{m}}$. Finally, we use the EC embedding and matched memory embedding to predict matching result. Meanwhile, we also optimize the distance between two embeddings to differentiate inclusion and exclusion ECs.
}
\label{fig:model structure}
\vskip -1em
\end{figure*}

\section{Problem Formulation}

Below we define the input data and the modeling problem in this paper.  We summarize the notations in Table.~\ref{table:notation}.

\begin{definition}[\textbf{Patient Records}]
In longitudinal EHR data, each patient can be represented as a sequence of multivariate observations $\bm{P} = [\bm{v}_{1}, \bm{v}_{2}, ..., \bm{v}_{T}]$, where $T$ is the number of visits. Each visit is represented by $\bm{v}_{t} = [\bm{d}_{t_1},\ldots, \bm{d}_{t_i}, \bm{o}_{t_1}, \ldots, \bm{o}_{t_j}, \bm{p}_{t_1}, \ldots, \bm{p}_{t_k}]$ with diagnosis $\bm{d}_{t_i} \in \mathcal{D}$, medication $\bm{o}_{t_j} \in \mathcal{O}$ and procedure $\bm{p}_{t_k} \in \mathcal{P}$.  Here $\mathcal{D}$, $\mathcal{O}$ and $\mathcal{P}$ are the sets of diseases, medications and procedures, respectively.
For  simplicity, we use $\bm{g}_{t}$ to represent any medical code ($\bm{d}_{t}$, $\bm{o}_{t}$, or $\bm{p}_{t}$) in $\bm{v}_{t}$.
Each patient also has demographic features $\bm{v}_{d}$ including gender and age. \end{definition}
\begin{definition}[\textbf{Clinical Trials}]
For each clinical trial, its protocol $\bm{C} = [\bm{c}_{1}^{I}, ..., \bm{c}_{M}^{I}, \bm{c}_{1}^{E}, ..., \bm{c}_{N}^{E}]$ comprises two types of eligibility criteria: inclusion criteria and exclusion criteria.  where $M$ ($N$) denotes the number of inclusion (exclusion) criteria in the trial, $\bm{c}_{i}^{I}$ ($\bm{c}_{i}^{E}$) denotes the $i$-th inclusion  (exclusion) criterion. Each criterion $\bm{c}$ can be represented as a sequence of words: $\bm{c} = [\bm{w}_{1}, \bm{w}_{2}, ..., \bm{w}_{W}]$, where $W$ is the number of words in $\bm{c}$.
\end{definition}
\begin{definition}[\textbf{Medical Taxonomy}]
The medical taxonomy $G$ expresses the hierarchy of various medical concepts in the form of a parent-child relationship including diagnosis taxonomy $G_{\mathcal{D}}$, procedure taxonomy $G_{\mathcal{P}}$ and medication taxonomy $G_{\mathcal{O}}$. The leaf nodes in a taxonomy $G$ are detailed medical concepts (e.g. \textit{Type 2 diabetes mellitus}), while parent nodes represent more general concepts (e.g. \textit{Endocrine, nutritional and metabolic diseases}). $G$ can be built by using well-organized taxonomy of medical concepts (e.g. ICD\footnote{https://www.cdc.gov/nchs/icd/icd10.htm}, CCS\footnote{https://www.hcup-us.ahrq.gov/toolssoftware/ccs10/ccs10.jsp} or USC\footnote{https://www.wolterskluwercdi.com/drug-data/uniform-system-of-classification-cross-usc/}). We use the notation $G$ to represent a general taxonomy for medical code $\bm{g}_{t}$.
\end{definition}

\begin{problem}[\textbf{Patient Criteria Matching}]
Given a patient's visit records $P$ and a set of inclusion or exclusion criteria, we formulate the patient-criteria matching task as a \textit{multi-class classification} problem, which is to classify the matching results between  patients and ECs into three categories: "match", "mismatch", and "unknown" based on the similarity between patient records and trial criteria: $\hat{\bm{y}}(\bm{c},\bm{P}) \in \{match, mismatch, unknown\}$.
\end{problem}

\begin{problem}[\textbf{Patient Trial Matching}]
Given a patient's visit records $P$ and a clinical trial $\bm{C}$ that comprises a set of inclusion or exclusion criteria, we consider a patient and a trial are a match only if  the patient matches (i.e., confirms) all inclusion criteria and mismatches (i.e., refutes) all exclusion criteria in the trial.
\end{problem}


\begin{table}[h!]
\centering
\caption{Notations used in \mname.}
\label{table:notation}
\resizebox{\columnwidth}{!}{
\begin{tabular}{c|l}
\toprule[1pt]
Notations & short explanation \\ 
\hline 
$\bm{P}$ & patient record sequence \\
$\bm{v}_{t}$ & multivariate visit record at the $t$-th visit \\
$\bm{v}_{d}$ & patient demographic features \\
$\bm{g}_{t}$ & a medical code in $\bm{v}_{t}$\\
\hline
$G_{\mathcal{D}}, G_{\mathcal{O}}, G_{\mathcal{P}}$ & taxonomy  for disease, medication and procedure\\
$\mathcal{D}, \mathcal{O}, \mathcal{P}$ & sets of diseases, medications, and procedures\\
\hline
$\bm{C}$ & a clinical trial \\
$\bm{c}_{i}^{I}, \bm{c}_{i}^{E}$ & the $i$-th inclusion or exclusion criterion \\
$\bm{w}_{i}$ & the $i$-th word in a text sequence \\
\hline
$\bm{e}$ & learned EC sentence embedding \\
$\bm{m}^{k}$ & the memory slot at level $k$ \\
$\widetilde{\bm{m}}$ & retrieved patient embedding \\
$\hat{\bm{y}}(\bm{c},\bm{P}), \bm{y}(\bm{c},\bm{P})$ & predicted/ ground-truth   for criteria $\bm{c}$ with patient $\bm{P}$\\
\bottomrule[1pt]
\end{tabular}}
\end{table}

\section{The \mname model}
This section presents our \mname model (Fig.~\ref{fig:model structure}). \mname is a pseudo-siamese network consists of two branches: 1) a convolutional neural network-based branch to learn trial eligibility criteria (ECs) embeddings and 2) a taxonomy guided memory network branch to learn embeddings for patient records (EHR). Then we use an dynamic alignment and matching module to generate final outputs. \mname will embed two modalities of data from ECs and EHR into a shared latent space. In particular, we use EC embeddings as query to read memories in the memory network. Finally, we use the retrieved EHR embeddings and EC embeddings to jointly predict matching scores $\hat{\bm{y}}_{t}$.

\subsection{Trial Eligibility Criteria Embedding}
Trial  eligibility criteria (ECs) describe the inclusion and exclusion criteria of clinical trials using unstructured text. To learn embeddings for trial ECs, previous works use GloVe \cite{Pennington14glove:global} or BERT \cite{devlin2018bert} with max-pooling to obtain a static embedding vector for each criterion. However, this is not a good way to capture detailed information in sentences such as numerical values or units. We hope the learned EC embeddings can retain sentence-level semantics while also capture important detailed information in sentences. To this end, we use convolutional neural networks (CNN) to learn EC embedding, as  previous works \cite{hu2014convolutional} shows that CNN can generate rich matching patterns at different levels for semantic matching tasks.

First, we use pretrained BERT to generate contextualized word embedding for each word of EC. Specifically, we applied the Clinical BERT \cite{alsentzer2019publicly} pretrained on PubMed\footnote{https://www.ncbi.nlm.nih.gov/pubmed} text resources and MIMIC-III \cite{johnson2016mimic} doctor notes. Concretely, given an EC sentence $\bm{c}=[\bm{w}_{1}, ...,\bm{w}_{N}]$, the contextualized embeddings $\widetilde{\bm{c}}$ are calculated as:
\begin{equation}
    \widetilde{\bm{c}} = [\widetilde{\bm{w}}_{1}, ..., \widetilde{\bm{w}}_{N}] = BERT([\bm{w}_{1}, ...,\bm{w}_{N}])
\label{eq.sentence_embedding}
\end{equation}
where $N$ denotes the number of words in the EC sentence and $\widetilde{\bm{w}}_{i}$ is a contextualized word embedding for $\bm{w}_{i}$. The final embedding for the EC sentence is the concatenation of these word embeddings.

Then we use multiple one-dimensional convolutional layers with different kernel sizes to capture semantics at different granularity level (four levels in our experiment) \cite{you2018end}. We feed $\widetilde{\bm{c}}$ into the one-dimensional convolutional layers and concatenate the outputs to generate a vector $\bm{x}$ as:
\begin{equation}
    \bm{x} = [Conv(\widetilde{\bm{c}}, k_{1}),  Conv(\widetilde{\bm{c}}, k_{2}), Conv(\widetilde{\bm{c}}, k_{3}), Conv(\widetilde{\bm{c}}, k_{4})]
\end{equation}
where $Conv$ denotes the one dimensional convolutional operator with kernel size $k_1$ to $k_4$. Then we feed $\bm{x}$ into convolutional highway layers, which has been widely used in many natural language processing tasks to extract text semantics \cite{you2018end,srivastava2015highway}. Particularly, the outputs of highway networks are calculated as:
\begin{equation}
\begin{split}
    \bm{u} &= \sigma(Conv(\bm{x}, k)) \\
    \bm{v} &= \bm{u} \cdot Conv(\bm{x}, k) + \bm{x} \cdot (1-\bm{u})
\end{split}
\label{eq.highway}
\end{equation}
where $\sigma$ denotes the sigmoid activation. We set the stride to 1 and use same padding to make sure the output dimension of $\bm{v}$ is the same as the input dimension of $\bm{x}$.

Finally, we use max-over-time pooling~\cite{kim2016character} to reduce the dimension and obtain the final embedding $\bm{e}$ for the EC sentence in current trial as $\bm{e} = MaxPool(\bm{v})$. Each dimension in $\bm{e}$ denotes a semantic feature captured by convolutional filters.
\subsection{Taxonomy Guided Patient Embedding}

Memory networks are powerful frameworks to process, store and retrieve information from sequential data~\cite{weston2014memory}.
To enable the patient-trial matching task, we also learn effective patient representation from patient data (e.g., EHR) by using memory network. In this work, we developed a hierarchical memory network to leverage medical taxonomy. This is achieved by allowing each memory slot to  memorize concept on a specific level of the taxonomy.

Formally, given a patient's EHR data, the $t$-th visit $\bm{v}_{t}$ is represented by three types of medical concepts: diagnosis, medications and procedures. We encode each type of code using a separate memory component. Since each type of concept can be divided into different levels, from a more general concept to a more specific one. In this work, we use the Uniform System of Classification (USC) taxonomy,  a therapeutic classification system, to divide each concept into four levels. To this end,  our patient embedding memory network  consists of three separate memory networks to store diagnosis, medications and procedures respectively. Each  memory network has four memory slots to store information from fine-grained to coarse levels. This can be represented as:
\begin{equation}
\begin{split}
    \mathbf{m} &= [\mathbf{m}_{\mathcal{D}}, \mathbf{m}_{\mathcal{O}}, \mathbf{m}_{\mathcal{P}}]  \\
    &= [\mathbf{m}_{\mathcal{D}}^{1}, ..., \mathbf{m}_{\mathcal{D}}^{4},\mathbf{m}_{\mathcal{O}}^{1}, ..., \mathbf{m}_{\mathcal{O}}^{4},\mathbf{m}_{\mathcal{P}}^{1}, ..., \mathbf{m}_{\mathcal{P}}^{4}]
\end{split}
\end{equation}
$\bm{m}_{x}^{1}$ is at the root level (most general), while 
$\bm{m}_{x}^{4}$ is the leaf level (most specific).
Each memory slot $\bm{m}_{x}^{k}$ is a vector initialized by zero. At each visit, the values in the memory slots will be updated. Note that we use four levels because of the specific medical taxonomies used in the experiment. The method can be easily generalized to taxonomy of different levels like \cite{gao2019camp, choi2017gram}. \\

\noindent\textbf{Augment medical codes with textual description}.
It is worth noting that we do not use the one-hot encoded medical codes as the input like previous trial matching and other clinical prediction works do~\cite{zhang2020deepenroll,choi2018mime, gao2020stagenet, ma2019adacare}.
Because, in clinical trial ECs, medical concepts are described in natural language rather than medical codes. 
Therefore, using textual description of medical codes to embed patients' EHR can help the model find matched concepts. All the textual description of each medical concept can be found in the Uniform System of Classification (USC). For example, we use the textual description \textit{Contact dermatitis and other eczema} for the diagnosis code 692.9. More specifically each medical concept can be represented as lists of words, i.e., $\bm{g}_{t} = [\bm{w}_{1}, ..., \bm{w_{L}}]$, where $L$ is the length of the text description for the code $\bm{g}_{t}$. We use BERT with max-pooling to generate the concept embedding:
\begin{equation}
    \widetilde{\bm{g}}_{t} = MaxPool(BERT([\bm{w}_{1}, ..., \bm{w}_{L}]))
    \label{eq.concept_embedding}
\end{equation}
Specifically, when the patient have multiple diagnosis / medications / procedures, we use max-pooling layer to aggregate the final embedding for each type of codes (i.e., we will obtain three code embeddings $\widetilde{\bm{d}}_{t}$, $\widetilde{\bm{o}}_{t}$ and $\widetilde{\bm{p}}_{t}$).\\

\noindent\textbf{Update memories at each visit}.
Since patients' EHR are sequential data, we update the memory slots at each visit. We adopt the \textit{erase}-followed-by-\textit{add} update mechanism in \cite{zhang2017dynamic, gao2019camp}. It allows us to erase unnecessary information at each visit and then add new information dynamically. Given the medical code embedding $\widetilde{\bm{g}}_{t}$ at $t$-th visit, we can obtain its parent code embeddings $\widetilde{\bm{g}}_{t}^{1}$ to $\widetilde{\bm{g}}_{t}^{3}$ in the corresponding taxonomy and $\widetilde{\bm{g}}_{t}$ is the leaf node $\widetilde{\bm{g}}_{t}^{4}$.     For the code at level $k$ ($\widetilde{\bm{g}}_{t}^{k}$) and corresponding memory slot ($\bm{m}_{G}^{k}$), we calculate the \textit{erase} and \textit{add} gate as:
\begin{equation}
    \textbf{erase}_{t} = \sigma(\mathbf{W}_{e}\widetilde{\bm{g}}_{t}^{k}+\mathbf{b}_{e}), \quad
    \textbf{add}_{t} = tanh(\mathbf{W}_{a}\widetilde{\bm{g}}_{t}^{k} + \mathbf{b}_{a})
\label{eq.erase}
\end{equation}

Then we update the memory in the slot as:
\begin{equation}
    \bm{m}_{G}^{k} \shortleftarrow \bm{m}_{G}^{k} \odot (1-\textbf{erase}_{t}) + \textbf{add}_{t}
\label{eq.update}
\end{equation}
where $\odot$ denotes the element-wise product.

\subsection{Attentional Record Alignment and Dynamic Criteria Level Match}
Previous patient trial matching works learn an alignment matrix which map patients' EHR embeddings and EC embeddings to the same latent space. However, each patient may enroll more than one trial and vice versa. Since each trial generally focus on certain diseases, the semantic distance between trials may be large. Trying to push the patient's EHR embedding to be close to different EC embeddings may confuse the embed function, or the patient embedding could be mapped to near average position of those EC embeddings. Such situation is similar to the multi-label image classification case in computer vision \cite{ren2015multi}.

To overcome this problem, we let each EC correspond to the sub-memories that match current trial best. Concretely, we use the inclusion criteria embedding $\bm{e}_{I}$ or exclusion criteria embedding $\bm{e}_{E}$ as queries to attentively visit the memory network. Given the memory slot $\bm{m}_{G}^{k}$, the attention weight for current slot $a_{k,G}$ is calculated as:
\begin{equation}
    a_{k,G} = \frac{exp({\bm{m}_{G}^{k}}^{\text{T}}MLP(\bm{e}))}{{\sum_{x \in {\{\mathcal{D},\mathcal{O},\mathcal{P}\}}}\sum_{i=1}^{4}{exp({\bm{m}^{i}_{x}}^{\text{T}}MLP(\bm{e}))}}}
\label{eq.attention}
\end{equation}
where $MLP(\cdot)$ is a multi-layer perceptron to align the dimension between $\bm{e}$ and $\bm{m}_{G}^{k}$. A large $a_{k,G}$ may indicate that the queried trial is highly related to the information stored in memory slot $\bm{m}_{G}^{k}$.

Given an criteria $\bm{e}$, we can obtain the best matching memories $\widetilde{\bm{m}}$ as:
\begin{equation}
\begin{split}
    \widetilde{\bm{m}} &= {\sum_{x \in {\{\mathcal{D},\mathcal{O},\mathcal{P}\}}}}\sum_{i=1}^{4}{a_{i,x}\bm{m}_{x}^{i}}
\end{split}
\label{eq.retrieve}
\end{equation}

Specifically, we also use MLP to learn a embedding for the patient's demographics as: $\bm{m}_{d} = MLP(\bm{v}_{d})$, where $\bm{m}_{d}$ has the same dimension with $\bm{\widetilde{m}}$. Finally, we concatenate the EHR embedding and the retrieved memory to predict whether the criteria $\bm{e}$ matches the patient as:
\begin{equation}
    \hat{\bm{y}} = softmax(\widetilde{\bm{m}} \oplus MLP(\bm{m}_{d} \oplus \bm{e}))
\label{eq.predict}
\end{equation}
where $\oplus$ denotes the concatenation operation and we use MLP to map the criteria and demographics embedding to the same embedding space with $\widetilde{\bm{m}}$.

\subsection{Joint learning with Explicit Inclusion/Exclusion Criteria Handling}

During model optimization, in order to maximize the patient-trial matching as well as to explicit handle the difference between inclusion and exclusion criteria, we designed a composite loss function with the following loss terms.\\

\noindent\textbf{Classification Loss}. We use a cross-entropy loss term in Eq.~\ref{eq.loss1} to  optimize the classification performance between prediction $\hat{\bm{y}}$ and ground truth $\bm{y}$:
\begin{equation}
    \mathcal{L}_{c} = -(\bm{y}^{\text{T}}log(\hat{\bm{y}}) + (1-\bm{y})^{\text{T}}log(1-\hat{\bm{y}}))
\label{eq.loss1}
\end{equation}

\noindent\textbf{Inclusion/Exclusion  Loss}. In addition, we also construct a loss term to explicitly handle the match between patient embedding and EC embedding for inclusion criteria and exclusion criteria. The loss term enables the model to extract different features (e.g. negation words) in the inclusion/exclusion criteria and thus help decide whether to include or exclude the patient.
Mathematically, this boils down to minimize the distance between the retrieved patient memory and the embedding for inclusion criteria (i.e., $(\bm{e_{I}}, \widetilde{\bm{m}}_{I})$) while maximize the distance between the memory and the embedding for exclusion criteria (i.e., $(\bm{e_{E}}, \widetilde{\bm{m}}_{E})$). We construct the loss term with the following pairwise distance loss:
\begin{equation}
    \mathcal{L}_{d} = 
    \left \{
    \begin{aligned}
        & 1-d(\bm{e}, \widetilde{\bm{m}}_{I})), & if\:\bm{e}\:is\:\bm{e}_{I}\\
        & max(0, d(\bm{e}, \widetilde{\bm{m}}_{E}) - \alpha), & if\:\bm{e}\:is\:\bm{e}_{E}
    \end{aligned}
    \right.
\label{eq.loss2}
\end{equation}
where $d(\cdot,\cdot)$ is the similarity function between two vectors. In this work we use the cosine similarity function to measure the distance between two modalities of data. Here $\alpha$ is a hyper-parameter which denotes the minimum distance between the embedding of exclusion criteria  and the patient memory. If a patient matches an inclusion criterion, the model will minimize the cosine distance between the two embeddings to make $1-d(\bm{e},\widetilde{\bm{m}}_{I})$ close to $0$. If the patient is excluded by an exclusion criterion, the distance between two embeddings (i.e. $max(0, d(\bm{e}, \widetilde{\bm{m}}_{E}) - \alpha)$) should be no smaller than $\alpha$. This allows $\bm{e}_{I}$ and $\bm{e}_{E}$ have different distance to $\widetilde{\bm{m}}$ in the latent embedding space.

Finally, we jointly minimize the loss functions by back propagation in an end-to-end way as:
\begin{equation}
    \mathcal{L} = \mathcal{L}_{c} + \mathcal{L}_{d}
\label{eq.loss}
\end{equation}

Our \mname algorithm  is summarized in Algorithm.~\ref{alg:Framwork}.

\begin{algorithm}[h!] 
\caption{The \mname model} 
\label{alg:Framwork} 
\begin{algorithmic}
\REQUIRE ~~\\ 
Patient records $\bm{P} = [\bm{v}_{1}, \bm{v}_{2}, ..., \bm{v}_{T}]$, a set of inclusion criteria or exclusion criteria sentences $[\bm{c}_{1},...,\bm{c}_{N}]$. \\
\ENSURE ~~\\ 
Initialize all memory slots $\bm{m}$ to zero.
\FOR{$i=1$ to $T$}
\STATE Generate all medical concept in $\bm{v}_{i}$ embedding using Eq. \ref{eq.concept_embedding}\\ 
\STATE Update corresponding memory slots using Eq. \ref{eq.update} and \ref{eq.erase}; \\
\ENDFOR \\
\FOR{$i=1$ to $N$}
\STATE Generate EC sentence embedding $\widetilde{\bm{c}}_{i}$; \\
\STATE Obtain the convolution results using Eq. \ref{eq.highway}; \\
\ENDFOR \\
Retrieve the best-matching memories $\widetilde{\bm{m}}$ using Eq. \ref{eq.retrieve} and \ref{eq.attention}; \\
Calculate matching results $\hat{\bm{y}}$ using Eq. \ref{eq.predict}; \\
Update parameters by optimizing the loss in Eq. \ref{eq.loss1}, \ref{eq.loss2} and \ref{eq.loss}.
\end{algorithmic}
\end{algorithm}

\section{Experiment}
We evaluate \mname by comparing against state-of-the-art baselines on a real-world patient trial matching dataset.  The code of \mname is publicly available at \footnote{https://github.com/v1xerunt/COMPOSE}.

\subsection{Experimental Setup}

\noindent\textbf{Dataset description}
We evaluated \mname using the data below.
\begin{enumerate}[leftmargin=*]
   \item \textbf{Clinical Trial Data} We randomly selected 590 clinical trials with varying disease domains from publicly available data source (ClinicalTrials.gov). We extract the inclusion criteria and exclusion criteria from these trials. In total, we obtain $12,445$ criteria-level (i.e., sentence-level) EC statements.
   \item \textbf{Patient EHR Data} We extract patient claims data from IQVIA's real-world patient database, which can be accessed by request. In total we have EHR records from $83,371$ patients from 2002 to 2018, where each patient is a match to at least one trial in the previous extracted trial dataset. The patient information is encoded in a longitudinal prescription and medical claims data including diagnosis, procedures and medications.
\end{enumerate}

We label each inclusion/exclusion criterion and their corresponding matched patients' EHR as "\textit{match}"/"\textit{mismatch}". For each criterion, we randomly sample one inclusion criterion and exclusion criterion from another trial and label them as "\textit{unknown}". In all we have 397,321 labelled pairs. The data statistics are  in Appendix.

\noindent\textbf{Baselines}
We evaluated \mname against the following baselines.
\begin{enumerate}[leftmargin=*]
    \item \textbf{LSTM+GloVe}~\cite{hochreiter1997long}. We use LSTM to learn the representation for longitudinal EHR data and use GloVe~\cite{Pennington14glove:global} followed by a max-pooling layer to learn the sentence embedding for ECs. Then we concatenate the outputs to predict the matching results.
    \item \textbf{LSTM+BERT}~\cite{devlin2018bert} We use LSTM to learn EHR embeddings and use BERT to learn EC sentence embeddings. Then we concatenate the outputs to predict the matching results.
    \item \textbf{Criteria2Query}~\cite{yuan2019criteria2query} consists of a systematic information extraction pipeline that uses basic text mining solution (such as named entity recognition) to parse unstructured text data and translate them to a set of structured attributes. And then use these attributes to identify patient cohorts.
    \item \textbf{DeepEnroll}~\cite{zhang2020deepenroll} is the previous state-of-the-art deep learning model for patient trial matching task. The structure of DeepEnroll is similar to \textbf{LSTM+BERT}, but it uses MiME~\cite{choi2018mime} instead of LSTM to learn EHR embedding. DeepEnroll uses an alignment matrix to predict the matching results.
\end{enumerate}
GloVe and BERT are common models for text-related retriveal tasks. We use them as our baseline pseudo-siamese structures. In addition to these baselines, we also perform ablation studies by
 comparing \mname against its reduced models:
\begin{enumerate}[leftmargin=*]
    \item \textbf{\mname-MN} We reduce the memory network from \mname. We use an LSTM model to learn the EHR embedding, and we use one-hot encoded codes instead of textual decriptions.
    \item \textbf{\mname-Highway} We reduce the highway network from \mname. We use a regular three-layer CNN to learn the trial embedding.
    \item \textbf{\mname-$\mathcal{L}_{d}$} We reduce the $\mathcal{L}_{d}$ loss term from \mname. We use just $\mathcal{L}_{c}$ to optimize the model. Such setting will make the model not explicitly differentiate inclusion criteria and exclusion criteria.
\end{enumerate}

\noindent\textbf{Evaluation strategy and metrics}
(1) \textbf{Criteria Level}: We use all labelled patient-criterion pairs to train and evaluate our model. We follow previous works~\cite{zhang2020deepenroll} to use the accuracy score (\textbf{Accuracy}), area under the receiver operating characteristic curve (\textbf{AUROC}), and area under the precision-recall curve (\textbf{AUPRC}) to evaluate model performance. Since the matching task is casted as a multi-label classification task, we calculate all scores using micro-average. 
(2) \textbf{Trial Level}: For trial level matching tasks, if patient A has enrolled in trial B, we first obtain all matching results between A and all criteria in B, then we aggregated the results to get the final matching results between patient A and trial B. The result is considered correct when the prediction results between A and all inclusion criteria in B are "\textit{match}" and the results between A and all exclusion criteria are "\textit{mismatch}". We use the accuracy score to evaluate the performance. In practice, some inclusion or exclusion criteria can be too strict to prevent finding patients, yet non-essential thus can be modified. We choose four threshold (0.7, 0.8, 0.9 and 1) to simulate that trial recruiters may loose the restriction to enroll enough patients. For example, threshold 0.7 means we consider a patient match the trial when the patient only matches 70\% criteria in the trial.

We fix a test set of 30\% patients, and divide the rest of the dataset into the training set and validation set with a proportion of 90\%:10\%. We fix the best model on the validation set and report the performance on the test set. We perform five random runs and report both mean and standard deviation for testing performance except for Criteria2Query, since it requires no training process.


\subsection{Results}

We designed experiments to answer the following question.

\noindent \textbf{Q1}. How does \mname perform in patient-trial matching?

\noindent \textbf{Q2}. How does \mname perform for various disease categories? 

\noindent \textbf{Q3}. How does \mname perform for trials at each phases?

\noindent\textbf{Q4}. Sensitivity of performance for different matching threshold.

\noindent\textbf{Q5}. Cases for showing the attentional record align mechanism.



\subsection*{Q1. Performance on Patient Trial Matching}

The results for trial-level matching are shown in Table \ref{tab:trial_performance}. The accuracy score is computed based on 100\% matching criteria (i.e. a patient matches a trial only when the patient matches all inclusion criteria and mismatches all exclusion criteria in the trial). \mname significantly outperfoms other state-of-the-art methods. Compared to the best baseline model DeepEnroll, \mname achieves 24.3\% relatively higher accuracy score compared with DeepEnroll and 36.3\% relatively higher accuracy score compared with Criteria2Query .

\begin{table}[h!]
    \caption{ Patient-Trial matching. Performance are measured by accuracy based on the match of $100\%$ criteria. }
    \label{tab:trial_performance}
\begin{tabular}{llc}
\toprule
                          & \textbf{Model}   & \textbf{Accuracy} \\ \hline
\multirow{5}{*}{Baselines} & LSTM+GloVe      & 0.4294$\pm$0.010             \\
                          & LSTM+BERT        & 0.5460$\pm$0.008        \\
                          & Criteria2Query   & 0.6147$\pm$-               \\
                          & DeepEnroll       &  0.6737$\pm$0.021  \\ \hline
\multirow{3}{*}{Reduced}   & \mname-MN       &  0.7833$\pm$0.011  \\
                          & \mname-Highway   &  0.8102$\pm$0.009       \\
                          & \mname-$\mathcal{L}_{d}$ &    0.8212$\pm$0.010  \\ \hline
Proposed                   & \mname           & \textbf{0.8373$\pm$0.012} \\
\bottomrule
\end{tabular}
\end{table}

Among all baselines, Criteria2Query and DeepEnroll achieves better performance. All reduced models of \mname also outperforms all baseline models. And the results also show that textual description enhanced code embeddings and the hierarchical memory network that performs dynamic matching contributes most to the performance. Below we also provide the patient criteria matching results in Table~\ref{ec-level}. Compared to the best baseline model DeepEnroll, \mname achieves 8.8\% relatively higher accuracy, 4.7\% higher AUROC and 3.3\% higher AUPRC. The strong criteria level performance of \mname provides a good foundation for patient-trial matching. 

\begin{table}[h!]
    \caption{Performance on Criteria Level matching. }
    \resizebox{1\columnwidth}{!}{
\begin{tabular}{llccc}
\toprule
                          & \textbf{Model}   & \textbf{Accuracy} & \textbf{AUROC} & \textbf{AUPRC} \\ \hline
\multirow{3}{*}{Baselines} & LSTM+GloVe             & 0.722$\pm$0.010 &0.789$\pm$0.009&0.784$\pm$0.009 \\
                          & LSTM+BERT        &0.834$\pm$0.008&0.845$\pm$0.007&0.840$\pm$0.007  \\
                          & DeepEnroll       &0.869$\pm$0.012&0.936$\pm$0.013&0.947$\pm$0.011 \\ \hline
\multirow{3}{*}{Reduced}   & \mname-MN       &      0.899$\pm$0.012          &    0.955$\pm$0.013            &    0.960$\pm$0.010           \\
                          & \mname-Highway        &  0.912$\pm$0.007       &   0.965$\pm$0.007    & 0.967$\pm$0.009                        \\
                          & \mname-$\mathcal{L}_{d}$ & 0.939$\pm$0.010  &   0.976$\pm$0.009          &      0.973$\pm$0.007                      \\ \hline
Proposed                   & \mname&\textbf{0.945$\pm$0.008}&\textbf{0.980$\pm$0.007}&\textbf{0.979$\pm$0.008} \\
\bottomrule
\end{tabular}}
\label{ec-level}
\end{table}

\noindent\textbf{Varying Length of Patient Record}
It is challenging for matching patients with longer records due to  gradient vanishing issues of deep learning models or evolving health conditions of patients. Here we  experimentally explore how \mname performs in matching trials with patients who have short or long records.
We categorize patients into three groups based on the length of their  EHR records: Short (1 visit), Medium (2-3 visits), Long ($\ge4$ visits). We report the patient-trial matching performance for each group in Table \ref{tab:len_performance}.  

\begin{table}[h!]
    \caption{Performance (measured by accuracy) on trial level matching for different length of records.}
    \label{tab:len_performance}
    \resizebox{0.8\columnwidth}{!}{
\begin{tabular}{lccc}
\toprule
                           \textbf{Model}   & \textbf{Short} & \textbf{Medium} & \textbf{Long} \\ \hline
                        LSTM+GloVe             &0.4906&0.4328&0.0000 \\
                         LSTM+BERT        &0.5484&0.5512&0.5338   \\
                          Criteria2Query  &0.6833&0.5989&0.5172 \\
                           DeepEnroll       &0.6779&0.6797&0.6443 \\ \hline
             \mname&\textbf{0.8420}&\textbf{0.8389 }&\textbf{0.8350} \\
\bottomrule
\end{tabular}}
\end{table}

From  Table \ref{tab:len_performance}, we can observe that for most models it is easier to match patients with  short and medium length of records to trials. This is probably due to  patients with shorter sequences tend to have simpler health conditions, while patients with longer records tend to have more complex condition or condition changes, which cause their EHR to have irrelevant information that confused the patient-trial matching model. Compared with baselines,  \mname  have robust performance for patients with different length of records, this is because \mname uses dynamic memory network to store patients' EHR information, which has better capability to reserve fine-grained information in different slots. 

\subsection*{Q2. Varying  Disease Types } 

We also conduct experiments to explore how our model performs on different types of diseases. We select trials related to  chronic diseases, oncology and rare diseases. Particularly, we consider  19 trials on  9 chronic diseases including chronic pain, chronic obstructive pulmonary disease, etc. For oncology trials, we consider 33 trials on 18 oncologies including gastric cancer, lung cancer, etc.  we also select 5 rare diseases related trials including Glioma, Polymyositis, etc. More details are provided in Appendix.

From the results in Table \ref{tab:cohort_performance}, \mname generally outperformed other baseline methods and  outperformed best baseline DeepEnroll by 77.3\% relative higher accuracy for chronic diseases. For cancer and rare diseases cohorts, baseline models fail to match correct patients. However, Criteria2Query outperforms most baselines because it requires no training process, therefore insufficient data does not effect its performance.

For  \mname, the performance for matching patients with trials designed for chronic diseases is  lower than the other two disease types. This is due to patients with chronic diseases often have complex condition and heterogeneous manifestation. So the criteria for these diseases often have general descriptions. 
For example, most trials in the data are related to chronic pain (47.4\%), which is a common symptom often caused by other diseases. The ECs for trials on chronic pains often contain vague description such as \textit{Adolescents experiencing chronic pain of any type} or \textit{Have a history of non-cancer pain in past 6 months}. It is difficult for automatic matching methods to match patient with such terms like\textit{any type} or \textit{non-cancer}, thus results in low accuracy. 

In contrast, trials on oncology and rare diseases have more strict criteria in recruiting patients.  Most of  these ECs requires different medications and diagnosis from different aspect. Therefore, the task is much more difficult than common matching tasks for baseline modes, since these models have to align the patient representation with  each criterion. However, there is little training and testing data available  to finish such a complex task for most baseline models. We will further discuss this in the case study section.

\begin{table}[h!]
    \caption{Performance (measured by accuracy) on trial level matching  for different disease types. }
    \label{tab:cohort_performance}
    \resizebox{\columnwidth}{!}{
\begin{tabular}{lccc}
\toprule
                          \textbf{Model}   & \textbf{Chronic Diseases} & \textbf{Oncology} & \textbf{Rare Diseases} \\ \hline
                            LSTM+GloVe            &0.1793&0.0000&0.0000 \\
                            LSTM+BERT        &0.2062&0.0000&0.0000   \\
                           Criteria2Query  &0.5103&0.2722&0.2292 \\
                            DeepEnroll       &0.3345&0.0000&0.0000 \\ \hline
              \mname&\textbf{0.5931}&\textbf{0.6370}&\textbf{0.6875} \\
\bottomrule
\end{tabular}}
\end{table}

\subsection*{Q3. Varying Trial Phases } 
From the results in Table \ref{tab:phase_performance} , it is easy to see \mname significantly outperforms other models on different phases. Compared to the best baseline models DeepEnroll and Criteria2Query, \mname achieves 155\% relative higher accuracy for phase I trials, 19\% higher accuracy for phase II trials and 27\% higher accuracy for phase III trials.

Compared all phases, the matching performance on Phase I trials is generally much lower. This is due to phase I trials are designed to test a new regimen's tolerability and toxicity. These trials usually enroll a limited number of patients who have exhausted other treatment options. Consequently in our training data we also have much less Phase I trials and fewer patients available for training and testing: only 5\% patients enrolled in phase I trials, while much more patients are enrolled in phase II (42\%) and III trials (53\%).

\begin{table}[h!]
    \caption{Performance (measured by accuracy) on trial level matching for different trial phases.}
    \label{tab:phase_performance}
    \resizebox{0.85\columnwidth}{!}{
\begin{tabular}{lccc}
\toprule
                           \textbf{Model}   & \textbf{Phase I} & \textbf{Phase II} & \textbf{Phase III} \\ \hline
                             LSTM+GloVe             &0.0008&0.5865&0.3743 \\
                            LSTM+BERT        &0.0025&0.6045&0.4862   \\
                           Criteria2Query  &0.3025&0.6433&0.5870 \\
                            DeepEnroll       &0.2034&0.7493&0.6329 \\ \hline
 \mname&\textbf{0.5189}&\textbf{0.8939}&\textbf{0.8005}\\
\bottomrule
\end{tabular}}
\label{performance_length}
\end{table}

\subsection*{Q4. Varying Threshold of Matching} 
In practice, some inclusion or exclusion criteria can be too strict to prevent finding patients, yet non-essential thus can be modified. In this section, we will examine how  trial matching accuracy will change under different matching thresholds. We also show \mname can provide insights for guiding the adjustment of criteria.

Table \ref{tab:threshold_performance} shows the performance of trial matching results under threshold 70\%, 80\% and 90\% (i.e., a pair of patient and trial is considered  matching  when the patient matches 70\%/80\%/90\% of criteria of the trial). Criteria2Query is not applicable for this analysis since it requires all criteria in a trial to match patients. For \mname, a more strict matching threshold results in 2\% lower accuracy score. While for baseline models, their performance drops 5\%\textasciitilde6\% when the threshold rises from 80\% to 90\%. When threshold rises from 90\% to 100\%, performance drops by more than 11\%. 

In order to explore which criteria cause the performance drop, we examine three criteria in \textit{Tanezumab for Diabetic Peripheral Neuropathy} trial (NCT01087203) : (1) Other types of diabetic neuropathies; (2) Clinically significant neurological diseases; (3) Clinically significant psychiatric diseases.
The three  criteria  are successfully matched by \mname, but wrongly matched by other baselines when  the threshold is increased from 0.8 to 1. These criteria describe general disease cohorts rather than a specific disease. For baseline models that use RNN to encode medical codes, it is difficult to align such abstract disease cohort description with detailed codes in patients' EHR records such as \textit{Diabetes mellitus due to underlying condition with diabetic polyneuropathy}. Thanks to the hierarchical memory networks to store taxonomies of medical concepts, \mname can easily align the criteria to either a detailed code or a more general parent concept.

\begin{table}[h!]
    \caption{Performance (accuracy)  for different thresholds. }
    \label{tab:threshold_performance}
    \resizebox{1\columnwidth}{!}{
\begin{tabular}{lccc}
\toprule
                          \textbf{Model}   & \textbf{70\% Matching} & \textbf{80\% Matching} & \textbf{90\% Matching} \\ \hline
                            LSTM+GloVe            & 0.6218              &0.5862 &0.5057  \\
                           LSTM+BERT        & 0.7231& 0.6861& 0.6238\\
                           DeepEnroll       &  0.8225  & 0.7963 & 0.7422 \\ \hline
 \mname           & \textbf{0.9334} & \textbf{0.9193} &  \textbf{0.8915} \\
\bottomrule
\end{tabular}}
\end{table}

\subsection*{Q5. Case Studies} 
\begin{figure}[h!]
\centering
\includegraphics[width=1\columnwidth]{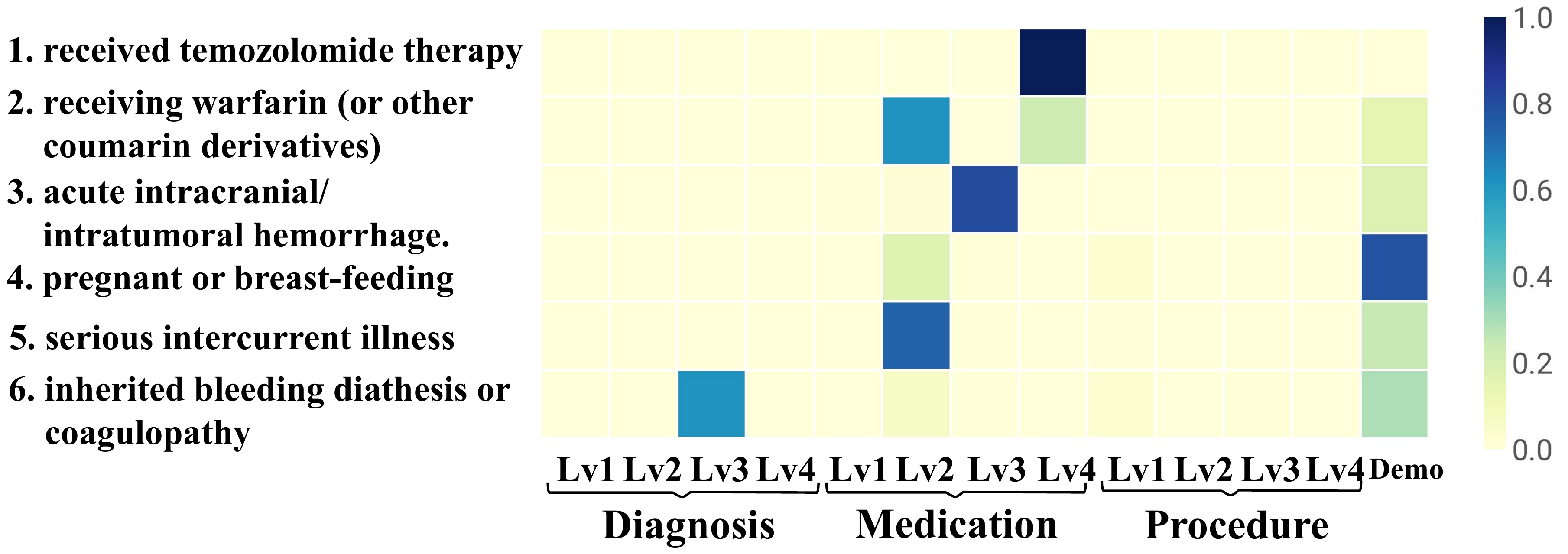}
\vskip -1em
\caption{Attention weights on the memory slots for the Cabozantinib trial for treating grade IV astrocytic tumors.}
\label{fig:case_att}
\vskip -1em
\end{figure}

To show how the attentional record align mechanism in \mname works, we choose a trial on Cabozantinib which treats grade IV astrocytic tumors. \mname successfully matches this trial (94\% matching) while all baselines fail (<50\% matching). Fig. \ref{fig:case_att} shows the attention weights on different memory slots for 6 selected criteria.

\begin{figure}[h!]
\vskip -1em
  \centering
  \subfigure[DeepEnroll]{
      \includegraphics[scale=0.31]{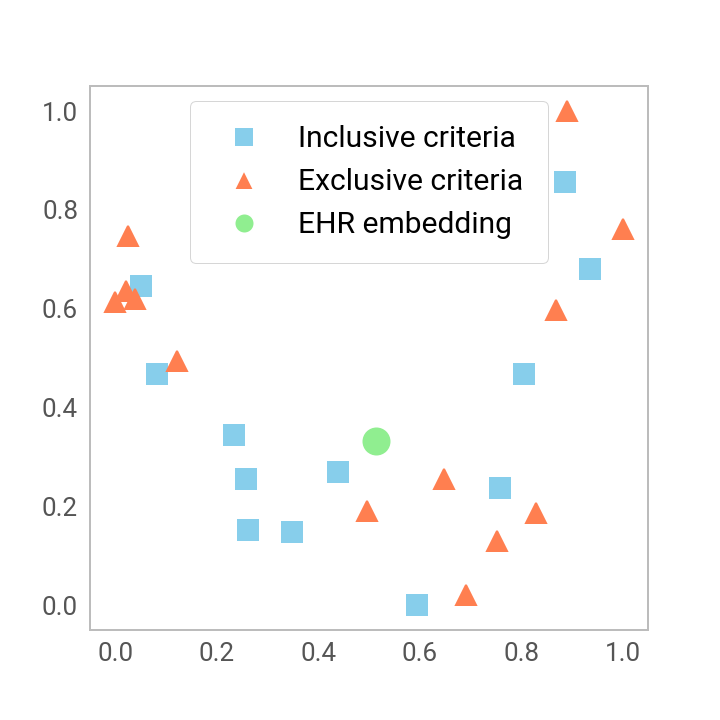}
      \label{fig:pca_lstm}
  }
  \subfigure[\mname]{
      \includegraphics[scale=0.31]{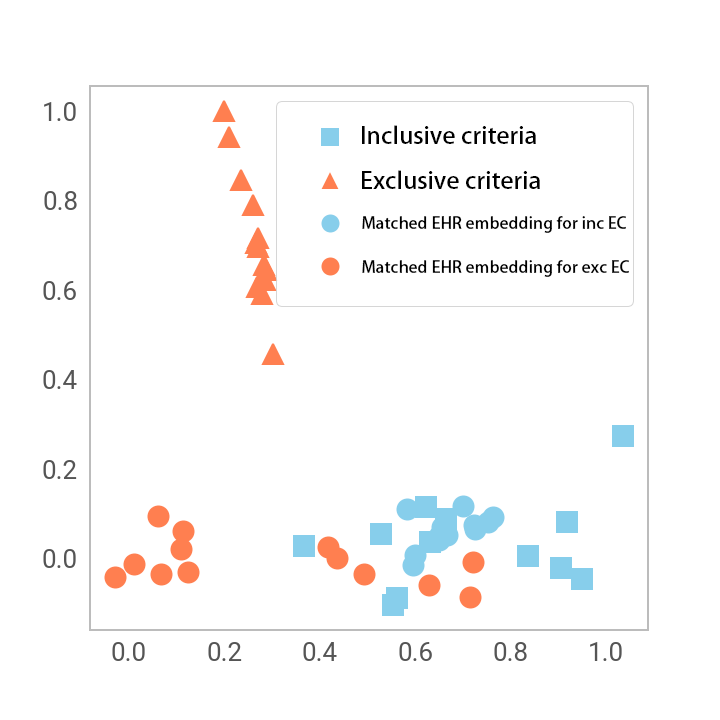}
      \label{fig:pca_deepenroll}
  }
  \vskip -1em
  \caption{Visualize EHR and EC embeddings using PCA}
  \label{fig:pca}
\vskip -1em
\end{figure}
For this trial, each criterion focuses on a single diagnosis or medication, so it is difficult for baseline models to match each criterion to a longer patient record. However, the attentional record align mechanism in \mname helps each criterion match the most related memory slots and therefore achieves dynamic matching. For example, the 4th criteria (\textit{pregnant or breast-feeding}) is aligned to the demographic slot. And the 2nd EC (\textit{receiving warfarin or other coumarin derivatives}) is aligned to the Lv.2 and Lv.4 of medication slots since \textit{warfarin} corresponds to a detailed medication while \textit{coumarin derivatives} corresponds to a higher level category. 

We also visually compared the learned  embeddings from \mname and DeepEnroll using principal component analysis (PCA) in Fig \ref{fig:pca}. For DeepEnroll, a patient (green circle) is aligned to multiple inclusion and exclusion criteria (squares and triangles), so the EC embeddings are mixed and lead to wrong prediction. However, for \mname, we use each EC as queries to match EHR records, so there are many EHR embeddings (blue and red circles) and each embedding is corresponding to a specific EC. So the model can have more accurate matching for different ECs. Besides, the inclusion and exclusion EC embeddings form different clusters, which means the model can differentiate them by optimizing the distance.

\begin{figure}[h!]
\centering
\includegraphics[width=0.8\columnwidth]{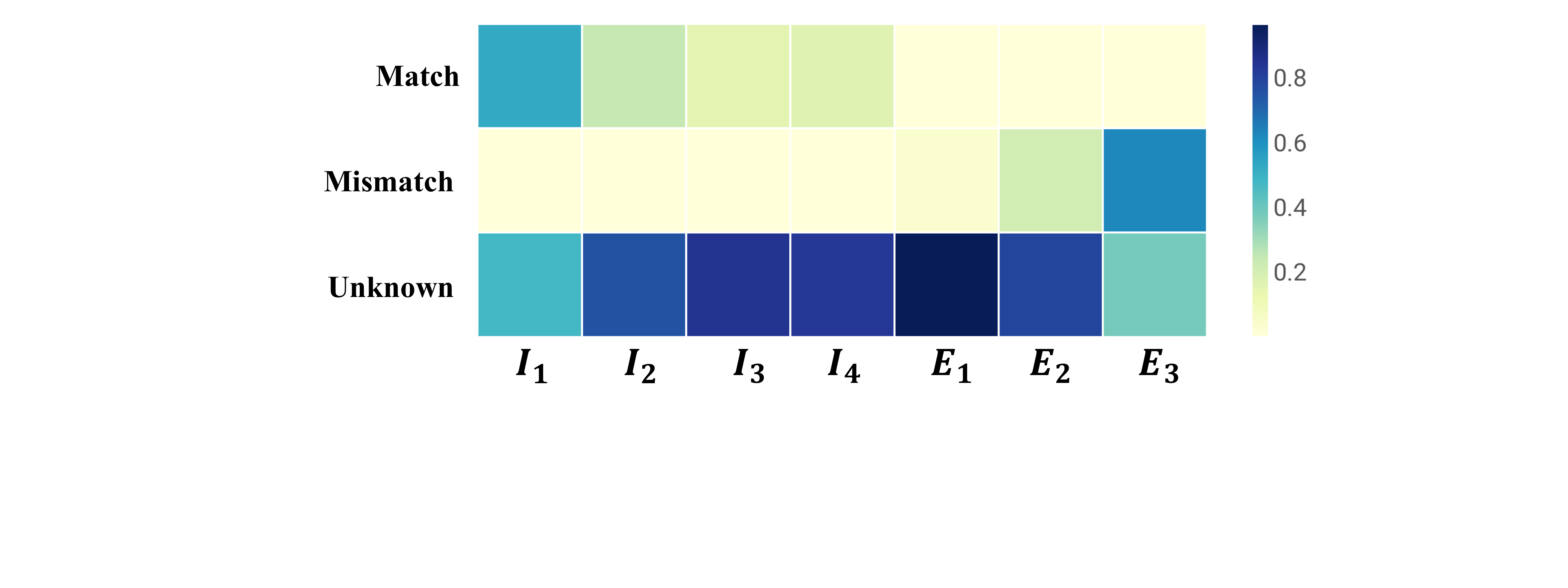}
\vskip -1em
\caption{Prediction results for NCT02998528 trial.  Inclusion criteria are denoted as $I$ and exclusion criteria as $E$.}
\label{fig:neg_case}
\vskip -1em
\end{figure}

We also found some trials difficult to find matching patients, e.g., the trial for \textit{Nivolumab Plus Ipilimumab or Nivolumab Plus Chemotherapy Versus Chemotherapy Alone in Early Stage Non-Small Cell Lung Cancer (NSCLC, NCT02998528)}. All models achieve lower than 50\% accuracy score for this trial. The ECs in this trial are listed in Appendix B, and we denotes inclusion criteria as $I$ and exclusion criteria as $E$. The prediction results of \mname for this trial and a case patient are shown in Fig \ref{fig:neg_case}.

The results show that \mname successfully matches $I_{1}$ and $E_{3}$ to the patient but classifies other ECs to \textit{unknown}. $E_{1}$ and $E_{2}$ describes a large group of diseases that models fail to match them to either a detailed code or a category. For $I_{2}$, $I_{3}$ and $I_{4}$,   EHR data is not enough to determine whether a patient matches these criteria.

\section{Conclusion}
In this work, we propose a cross-modal pseudo-siamese network model, \mname, to conduct patient-trial matching. \mname can perform  dynamic patient-trial matching based on learning taxonomy guided multi-granularity medical concept embedding. \mname is also augmented by a composite loss term to maximize the similarity between patient records and inclusion criteria while minimize the similarity between patient records and exclusion criteria. Experiments on real-world datasets demonstrated that \mname significantly outperforms state-of-the-art baselines.

\section{Acknowledgments}
This work is part supported by National Science Foundation award IIS-1418511, CCF-1533768 and IIS-1838042, the National Institute of Health award NIH R01 1R01NS107291-01 and R56HL138415.

\bibliographystyle{format}
\bibliography{references}

\appendix
\section{Details of Trial Data}
The basic statistics of the trial dataset and EHR dataset are shown in Table \ref{tab:stat}. The detailed diseases in different cohorts and the number of trials and patients for each disease are shown in Table \ref{tab:cohort}.
\begin{table}[h]
    \centering
        \caption{Data statistics}
    \label{tab:stat}
    \begin{tabular}{lc}
    \toprule
        \textbf{Statistic} &  \\
        \midrule
        \# of trials & 590\\
        Avg. \# of Inclusion criteria per trial & 6.67\\
        Avg. \# of Exclusion criteria per trial & 10.56\\
        Avg. criteria sentence length & 31.80\\
        \midrule
        \# of patients & 83,371\\
        \# of unique medications & 286\\
        \# of unique diagnosis & 421\\
        \midrule
        \# of labeled pairs & 397,321\\
        Avg. \# of trials per patient & 1.20\\
        Avg. \# of patients per trial & 169.49\\
    \bottomrule
    \end{tabular}
\end{table}

\begin{table}[h]
    \centering
    \caption{Diseases in different cohorts}
    \resizebox{1\columnwidth}{!}{
    \begin{tabular}{llcc}
        \toprule[0.7pt]
        Cohort & Disease & \# of trials & \# of patients\\
        \hline
         \multirow{5}{*}{Rare}&Polymyositis&1&11\\
         &Glioblastoma&1&16\\
         &Diffuse Large B-Cell Lymphoma&1&2\\
         &Glioma&1&2\\
         &Gastrointestinal Stromal Tumors&1&1\\
        \hline
        \multirow{18}{*}{Oncology} &Brain and Central Nervous System Tumors&1&60  \\
        &Unspecified Adult Solid Tumor&1&1 \\
        &Astrocytic Tumors&1&19\\
        &Advanced Solid Tumor&7&9\\
        &Ovarian Cancer&1&5\\
        &cMET-dysregulated Advanced Solid Tumors&1&1\\
        &Triple Negative Breast Cancer&1&2\\
        &Cancer&1&1\\
        &Non Small Cell Lung Cancer&6&6\\
        &Metastatic Colorectal Cancer&1&1\\
        &Colorectal Cancer&2&4\\
        &Tumors&2&2\\
        &Solid Tumor&3&3\\
        &PIK3CA Mutated Advanced Solid Tumors&1&1\\
        &mCRPC&1&2\\
        &Advanced or Metastatic Breast Cancer&1&2\\
        &Gastrointestinal Stromal Tumors&1&1\\
        &Advanced Gastric Cancer&1&1\\
        \hline
        \multirow{12}{*}{Chronic} &Chronic Demyelinating Polyradiculoneuropathy&1&56  \\
        &Chronic Low Back Pain&2&43 \\
        &Chronic Cluster Headache&3&49\\
        &Chronic Pain&4&20\\
        &Chronic Sinusitis With or Without Nasal Polyps&1&13\\
        &Chronic Myeloid Leukemia&3&5\\
        &Hepatitis C, Chronic&3&5\\
        &Chronic Severe Plaque-type Psoriasis&1&45\\
        &Treatment for Prevention of Chronic Migraine&1&21\\
        \bottomrule[0.7pt]
    \end{tabular}}
    \label{tab:cohort}
\vskip -1em
\end{table}

\section{Inclusion and Exclusion Criteria for NCT02998528 Trial}
All criteria in NCT02998528 trial is shown below. $I$ denotes the inclusion criteria and $E$ denotes the exclusion criteria.
\begin{itemize}
    \item $I_{1}$: Early stage IB-IIIA, operable non-small cell lung cancer, confirmed in tissue
    \item $I_{2}$: Lung function capacity capable of tolerating the proposed lung surgery
    \item $I_{3}$: Eastern Cooperative Oncology Group (ECOG) Performance Status of 0-1
    \item $I_{4}$: Available tissue of primary lung tumor
    \item $E_{1}$: Presence of locally advanced, inoperable or metastatic disease
    \item $E_{2}$: Participants with active, known or suspected autoimmune disease
    \item $E_{3}$: Prior treatment with any drug that targets T cell co-stimulations pathways (such as checkpoint inhibitors)
\end{itemize}

\section{Implementation details}
All methods are implemented in PyTorch~\cite{paszke2017automatic} and trained on an Ubuntu 16.04 with 64GB memory and a Tesla V100 GPU. We use Adam optimizer \cite{kingma2014adam} with a learning rate of 0.001. We use a batch size of 512 and train all model for 20 epochs. 

For hyper-parameter settings of each baseline model, our principle is as follows: For some hyper-parameter, we will use the recommended setting if it is available in the original paper. Otherwise, we determine its value by grid search on the validation set.
 
\begin{itemize}
    \item \textbf{LSTM+GloVe}. We use 300 dimensional GloVe embedding~\cite{Pennington14glove:global} as word embedding and max-pooling as sentence embedding method. Out-of-vocabulary (OOV) words are hashed to one of 100 random embedding each initialized to mean 0 and standard deviation 1. The hidden units of LSTM cell are set to 256.
    \item \textbf{LSTM+BERT}. The hidden units of LSTM cell are set to 512. We use the same pretrained clinicalBERT model in \mname.
    \item $\textbf{Criteria2Query}$. We directly use the existing model to produce the matching results for given trials and patients. This baseline does not require any training process or hyper-parameters.
    \item \textbf{DeepEnroll}. We use the recommended settings in the paper. However, to be fair, we use the same BERT model for DeepEnroll and our model.
    \item \textbf{\mname}. We set the dimension of all convolutional layers to 128. The kernel size $k_{1}$ to $k_{4}$ is set to 1, 3, 5, 7, and we use kernel size 3 for the highway layers. The hidden units of memory slots are set to 320. The margin $\alpha$ in loss term $\mathcal{L}_{d}$ is set to 0.3.
\end{itemize}

\end{document}